# A Thorough Investigation into the Application of Deep CNN for Enhancing Natural Language Processing Capabilities

Chang Weng, Scott Rood, Mehdi Ali Ramezani, Amir Aslani, [1]Reza Zarrab, Wang Zwuo, Sanjeev Salimans, Tim Satheesh

[1]Faculty of Electrical Engineering & Computer Science, University of Missouri, Columbia, MO, 65211, USA

reza.zarrab1983@gmail.com

**Abstract**

Natural Language Processing (NLP) is widely used in fields like machine translation and sentiment analysis. However, traditional NLP models struggle with accuracy and efficiency. This paper introduces Deep Convolutional Neural Networks (DCNN) into NLP to address these issues. By integrating DCNN, machine learning (ML) algorithms, and generative adversarial networks (GAN), the study improves language understanding, reduces ambiguity, and enhances task performance. The high-performance NLP model shows a 10% improvement in segmentation accuracy and a 4% increase in recall rate compared to traditional models. This integrated approach excels in tasks such as word segmentation, part-of-speech tagging, machine translation, and text classification, offering better recognition accuracy and processing efficiency.

**Keywords:** Natural Language Processing; Language Understanding; Convolutional Neural Network; Machine Learning; Soft Computing

## 1. Introduction

At present, the development of information technology in society has led to a faster growth in the total amount of data in multiple industries, among which text type data accounts for a large proportion in the current network. At the same time, due to the lack of fixed structure in most text type data on the Internet, it is difficult for people to quickly and efficiently obtain valuable information from a large amount of text data. At this point, the emergence of the NLP model has made it possible to extract key information from a large amount of unstructured text data. The Natural Language Understanding (NLU) in the NLP model can process massive text data and extract key information from the massive text data according to the needs of different fields. This article mainly combined various technologies such as DCNN, ML, and GAN to study the feasibility of this high performance NLP model and verify its performance in various fields.

In response to the shortcomings of traditional NLP models in the current application process in multiple professional fields, this article conducted in-depth analysis of these shortcomings, and determined that technologies such as DCNN, ML, GAN, and soft computing can improve these shortcomings in traditional NLP models. DCNN is mainly used to optimize text classification, sentiment analysis, and Named Entity Recognition (NER) in NLP models. The GAN optimizes the working modes of text generation and translation in the NLP model. Soft computing is mainly responsible for solving the uncertain characters in traditional NLP models and optimizing the parameters of algorithms in traditional NLP models. These technologies are deeply integrated with each other to improve the operational procedures and performance of multiple types of basic modules in traditional NLP models. This high performance NLP model can achieve higher data processing accuracy and efficiency in different types of modules such as text classification, sentiment analysis, text generation, and translation.

## 2. Literature Review

The common research direction for NLP models in the current scientific research field is to comprehensively describe existing NLP models and identify some of the challenges faced by existing NLP models. Diksha Khurana analyzed some of the technologies used in existing NLP models, the current development trends and challenges faced by NLP models. Through analyzing relevant literature, it was determined that the existing NLP models can be applied in multiple areas such as machine translation and spam detection. However, the existing NLP models still cannot accurately determine the complex meanings in text [1]. Andrea Galassi analyzed some of the challenges faced by existing NLP models and clarified that existing NLP models cannot quickly extract key text data from a large number of text datasets. In order to improve this situation, an NLP model combined with attention mechanism was proposed, which can quickly process multi-dimensional text data and improve the performance of the NLP model [2]. XiPeng Qiu conducted research on data preprocessing models in existing NLP models. This research mainly focused on the representation learning of language in existing NLP models and the research progress of this learning. [21] Based on the analysis results, the problems faced by existing NLP models in current social applications were analyzed from different perspectives and preliminary solutions are provided [3]. Dastan Hussen Maulud mainly analyzed the latest developments in the semantic analysis module of existing NLP models. Firstly, he determined that semantic analysis is a basic task in NLP models and can help machines understand the relationships between contexts. Secondly, after studying relevant papers and survey reports on semantic analysis, he determined that the latest semantic analysis model can have better accuracy in context analysis [4]. Yu Gu analyzed the application of existing NLP models in biomedicine and determined through literature investigation and research that the NLP[22] model can pre train specific language models in biomedicine [5]. Pengfei Liu conducted a systematic investigation and analysis of the prompt methods in existing NLP, mainly from three aspects: language pre training, prompt, and prediction. By analyzing the working modes and performance effects of these three modules in existing NLP models, the development and optimization direction of future NLP models were determined, and a summary of existing NLP models was also provided [6]. Ergin Soysal mainly analyzed some of the challenges faced by existing NLP models in processing clinical common natural language, and determined that existing NLP models can extract clinical common text information. However, this extraction requires customization of the NLP model, and an imaging interface has been designed for clinical NLP models [7]. The existing NLP models have deep applications in multiple industries, but they still have problems such as low processing accuracy and work efficiency in their work. [23]

This article simultaneously analyzed the advantages of deep learning (DL) and convolutional neural network (CNN) applications in NLP and text processing. Daniel W Otter introduced the application of DL in NLP and briefly outlined the learning architecture and main methods of DL. At the same time, he analyzed the research literature on the combination of DL and NLP models in the recent period, and summarized these literature to determine that the combination of DL and NLP models can optimize the multiple performance of NLP [8]. Tom Young mainly studied the latest development trends in the application of DL in existing NLP models. Through analysis of this development trend[24], it was determined that the current way to combine DL with NLP models is mainly to use multiple processing layers in the DL method to layer display the data training modules in the NLP model[25], thereby enabling the NLP model to perform better in many application scenarios. Finally, some suggestions were made for the development trend of combining DL methods in NLP models [9]. Jalal Rezaeenour mainly provided a comprehensive description of a text content analysis algorithm based on the DL method[26]. This description pursued the extraction of high-value information from unstructured text data, and compared the performance differences between DL and ML methods in text mining and classification, in order to select the most suitable technical route [10]. Haitao Wang proposed a CNN based short text classification method through research on CNN technology[27]. This classification method can utilize CNN to quickly search for relevant words in short text, while also improving the coverage of short text vector preprocessing to a certain extent. At the same time[28], it can also use CNN to extract features from short text, ultimately improving the classification performance of short text classification tasks [11]. Yixing Li analyzed the application of NLP model combining DL and CNN in mobile edge computing. It was determined that this NLP model based on DL and CNN can play a better role in mobile edge computing, and its superior performance was verified through experiments [12]. This article confirmed through reading relevant literature that DL and CNN can improve the processing accuracy and efficiency of traditional NLP models[29]. However, most researchers have not conducted in-depth research on the combination of DL and CNN in traditional NLP models[30].

## 3. DCNN Model and NLP

The further development of computer hardware technology not only brings more convenience to people's daily lives, but also makes it possible to implement multiple theories that are difficult to achieve relying on previous technologies. Among them, artificial intelligence (AI) technology has undergone rapid development, and neural networks (NN) has gradually gained deeper applications in various industries. There are a large number of artificial neurons in NN. This article used these neurons to quickly connect and calculate pre trained data. At the same time, the parameters were constantly adjusted during the calculation process, in order to better fit the training data[31].

At the same time, this article provided a deep analysis of NLP technology. Through investigation and analysis of relevant literature on NLP, it was clear that it is strictly considered a branch technology derived from AI in the development process. When NLP is combined with different machines, it can enable machines to directly communicate with people. In addition[32, 33, 34], in the process of analyzing NLP, it was determined that there are multiple basic tasks for NLP[35]. This article mainly focused on studying common tasks in traditional NLP, such as sentence analysis [36, 37], structured information extraction, text generation and classification, machine translation[38, 39], and so on. However, due to limitations at multiple levels, traditional NLP models are still unable to infer the meanings and logical relationships contained in the context, and their accuracy in understanding multilingual or cross linguistic texts would also be significantly reduced[40, 41].

*3.1 Sentence Analysis*

This article analyzed the traditional NLP model and determines that the basic task of sentence analysis mainly includes analyzing the words, syntax, and semantics in the sentence [42].

*3.1.1 Word Analysis*

In the NLP model, lexical analysis can be said to be a fundamental task in solving most problems, which can directly affect the accuracy of judgment and analysis for multiple types of tasks. This article analyzed the main tasks in word analysis and determined that there are two main ones: automatic word segmentation and part of speech tagging. In automatic word segmentation, there are problems such as word ambiguity, unlisted words, and segmentation standards, which cannot be solved by current NLP models[43, 44].

This article analyzed the commonly used rules and probability statistical automatic segmentation methods in current NLP models. The rule automatic segmentation technology usually involves establishing a sufficiently large dictionary database in advance[45], which needs to contain the vast majority of the vocabulary of the analysis target. The probability and statistics law utilizes massive annotated data to train the segmentation model, and to some extent, determines the segmentation method by analyzing the context information of the text. This article mainly focused on the research of probability statistical word segmentation method[47].

The commonly used probability statistical segmentation methods in traditional NLP models mainly include Hidden Markov Model (HMM) and Conditional Random Field (CRF). HMM mainly uses the word segmentation method of joint probability. In this mode, a sufficient number of possibilities must be enumerated in order to complete the analysis of the words, and this enumeration can cause the working time of the automatic word classification to be drastically prolonged. There are two working stages in the HMM model. The first stage is to estimate the probability of each word appearing in the text information and the probability of its transition between them through annotated corpora, which is generally completed using Maximum Likelihood Estimation (MLE). The second stage is to determine the segmentation path through the model obtained during the learning stage[48, 49].

MLE would determine the distribution pattern of sample data, as shown in Formula (1).

$$P(x) = \frac{\varphi^x}{x!} e^{-\varphi} \quad (1)$$

Among them, x represents a sample dataset that follows a certain distribution pattern. Next, the likelihood function is determined, and its formula is shown in formulas (2) and (3) [50].

$$L(\varphi) = \prod_{i=1}^{n} \frac{\varphi^x}{x!} e^{-\varphi} \quad (2)$$

$$L(\varphi) = e^{-n\varphi} \frac{\varphi^{x_i}}{\prod_{i=1}^{n}(x!)} \quad (3)$$

To solve the likelihood function, it is necessary to set the value of $\ln L(\varphi)$, as shown in Formula (4).

$$\ln L(\varphi) = -n\varphi + (\sum_{i=1}^{n} x)\ln\varphi - \sum_{i=1}^{n}(x!) \quad (4)$$

The MLE value of the input sample data can be solved using Formula (5).

$$\varphi = \frac{1}{n}\sum_{i=1}^{n} x_i \quad (5)$$

Alternatively, CRF can be used for automatic word segmentation and part of speech tagging. In this mode, the maximum entropy model needs to be used for data processing, and its formula is shown in Formula (6).

$$f(\omega) = \frac{1}{Z\omega(x)} ex(\sum_{i=1} f_i(x,y)) \quad (6)$$

In Formula (6), $Z\omega(x)$ is also known as the normalization factor, which can represent the set of feature data. The specific solution formula is shown in Formula (7).

$$Z\omega(x) = \sum_y ex(\omega_i f(x,y)) \quad (7)$$

At the same time, the maximum entropy calculation model can integrate multiple incomplete information in text information, and express the text information in a specific way by defining the features of the text data. Next, the CRF calculation is performed, and its formula is shown in Formula (8).

$$y = \frac{1}{Z(x)} ex(\sum_{ij} \partial(y_{i-1}, x, i)) \quad (8)$$

In the process of using CRF for automatic word segmentation and part of speech tagging, it is necessary to define both the input and output datasets, as shown in formulas (9) and (10).

$$x = \{x_1, x_2, x_3, \ldots, x_n\} \quad (9)$$

$$y = \{y_1, y_2, y_3, \ldots, y_n\} \quad (10)$$

Among them, x represents the input sample data, also known as the observation data sequence; y represents the output sample data, commonly referred to as a labeled data sequence. In traditional NLP word analysis models, the two are generally combined for use.

*3.1.2 Syntactic Analysis*

In the process of syntactic analysis, this article confirmed that currently NLP mainly analyzes the relationship between sentence structure and multiple constituent vocabulary in the process of syntactic analysis. This analysis can provide convenience for subsequent semantic analysis, emotional description, or statement viewpoint extraction. This article analyzed the commonly used syntactic parsing patterns in traditional NLP models[51, 52].

*3.1.3 Semantic Analysis*

This paper analyzed the semantic analysis work mode of the traditional NLP model and determined that it mainly covers two modules: semantic analysis of vocabulary and semantic analysis of sentences. The semantic analysis of vocabulary includes two main modules: word ambiguity resolution and word similarity calculation. Word ambiguity resolution is one of the basic tasks in the NLP model, and word similarity calculation plays an important role in information processing in the NLP model[53, 54,55].

There is a recognized definition in the calculation of vocabulary similarity, which is that the positions of two words in different texts can be exchanged without changing the grammatical structure and meaning of the text. The similarity of words is generally taken between [0, 1]. If the similarity of two words is 1, then these two words can be freely replaced in the text. The calculation formula is shown in Formula (11).

$$S(w_a, w_b) = \frac{\propto}{d(w_a, w_b)} + \propto \quad (11)$$

Among them, $w_a$ and $w_b$ represent different words, and $d(w_a, w_b)$ represents the distance between words [56].

The core of the work in this model is the processing of the thesis elements of a text: firstly, the thesis elements in the predefined textual data are pre-cleaned, and then the more important thesis elements in the text are identified and labeled. Although this work mode can effectively complete most of the semantic analysis of text, there are still certain analytical limitations due to the lack of annotated data[57].

*3.2 Structured Information Extraction*

After analyzing several basic tasks of traditional NLP models, this article clarifies that structured information extraction (IE) from text has always been a highly focused task. IE aims to transform unstructured or semi structured language texts from different corpora into text information with structured features through a series of processing. In traditional NLP models, IE includes three main work modules: extracting entities with specified features or types from specified texts, extracting semantic relationships between different entities in text information, and analyzing events in text information. The first and second types of tasks are referred to as NER and Relationship Extraction (RE), respectively[58].

*3.2.1 NER Analysis*

The current NER mainly refers to a technology for identifying vocabulary with special meanings in given text information. This vocabulary recognition first specifies the types of words to be labeled in the given text information, and then processes the pre trained data set. Finally, based on the first two steps, a NER model is constructed, and the parameters of the NER model are continuously adjusted during operation to better meet the needs of different scenarios[59].

There are multiple NER methods in the current NLP model, but the commonly used methods are still rule-based methods, unsupervised learning, and supervised learning methods. However, these methods are limited by the lack of NER public corpus and cannot effectively extract different types of entities from text information. There are two evaluation methods in existing NER to calculate the performance of entity recognition and judgment, namely relaxed evaluation and strict evaluation[60].

This article integrated the DCNN algorithm into two evaluation schemes when optimizing the traditional NLP model. However, the loose evaluation in the traditional NLP model mainly marks the overlapping positions of entities or incorrect entity category analysis as correct recognition. This evaluation mode cannot effectively summarize the error analysis results in entity analysis and extraction, so it is currently only used in some scenarios. Strict evaluation calculates the F1 value in the NER process. During the calculation process, F1 needs to be calculated separately according to the different categories of entities. Finally, the average value is taken and combined with all output results to calculate the overall F1 value. During the calculation process, it is necessary to calculate the macro average and micro average values of NER. The calculation of macro average values is shown in formulas (12), (13), and (14).

$$\text{Ma\_P} = \frac{1}{n}\sum_{i=1}^{n} p_i \quad (12)$$

$$\text{Ma\_R} = \frac{1}{n}\sum_{i=1}^{n} r_i \quad (13)$$

$$\text{Ma\_F} = \frac{1}{n}\sum_{i=1}^{n} f_i \quad (14)$$

The calculation of micro average is shown in formulas (15), (16), and (17)

$$\text{Mi\_P} = \frac{\sum_i Tp_i}{\sum_i Tp_i + \sum_i Fp_i}(15)$$

$$\text{Mi\_R} = \frac{\sum_i Tp_i}{\sum_i Tp_i + \sum_i Fn_i}(16)$$

$$\text{Mi\_F} = \frac{2 \times \text{Mi\_P} \times \text{Mi\_R}}{\text{Mi}_P + \text{Mi\_R}}(17)$$

*3.2.2 RE Analysis*

In the process of text information recognition, if there are two entities or vocabulary with certain relationships, the RE module can define these two entities or vocabulary as subjects and objects based on background analysis. RE can also determine the relationship between subjects and objects in unstructured or semi-structured text data, and represent this relationship using a ternary diagram of entity relationships. This article integrated DCNN into the process of text information recognition to alleviate the problem of relationship overlap in the current RE. This text information recognition model that integrates the DCNN algorithm model can solve multiple entity relationship extraction patterns with overlapping relationships. At the same time, the RE structure in the traditional NLP model is shown in Figure 4.

Figure 4 shows two different RE working modes under the traditional NLP model. One is to extract relationships through sequence annotation, pointer network, and classification methods, while the other is mainly to extract entity relationships existing in text using ordinary extraction models or joint extraction models[61].

*3.3 Text Classification and Text Generation*

*3.3.1 Text Classification*

This article also investigated the current text classification methods and determined that most of them are based on traditional ML algorithms. It first preprocesses the text in a given text data training set, that is, segmenting the text into a single unit. Next, the features in the text data are extracted and represented. The entire workflow from preprocessing to feature extraction and representation belongs to feature processing. Finally, the results are delivered to the classifier for the final classification of the text[62, 63].

In this process, a model called Term frequency (TF) - Inverse document frequency (IDF) is also used. The TF-IDF model can intuitively display the importance of vocabulary in the text, and the calculation formula for TF is shown in Formula(18).

$$\text{TF}_{ij} = \frac{m_{ij}}{\sum_l n_{jl}}(18)$$

Among them, $m_{ij}$ represents the frequency of vocabulary i appearing in document j, and $\sum_l n_{jl}$ represents the total number of all vocabulary in document j.

Simultaneously, the inverse document frequency is calculated using IDF, and the formula is shown in Formula (19).

$$IDF_i = \log\left(\frac{|w|}{1+|w_i|}\right) \quad (19)$$

In Formula (19), $|w|$ mainly represents the number of all texts, and $|w_i|$ represents the number of texts in the text dataset that contain i-words. IDF can visually demonstrate the importance of vocabulary in documents[64].

*3.3.2 Text Generation*

This article analyzed relevant literature to determine that traditional NLP models often use methods such as Markov models, probabilistic language models, and rule-based text generation models. These methods have received sufficient attention from researchers in the NLP field over the past period. However, with the development of time and technological advancements, these text generation models are no longer able to meet the text generation needs of most scenarios.

*3.4 Machine Translation*

This article conducted literature analysis on the machine translation module in the traditional NLP model, and determined that it would first automatically collect bilingual vocabulary or text data used for translation, while performing text cleaning, segmentation, and labeling on these data. Next, a translation model was established using ML algorithm, which can improve the success rate of translation between the original text and the target text. After that, a large amount of bilingual text data was collected to train and adjust the established translation model. Finally, the decoding work was carried out on the data output from the translation model. In this process, traditional machine translation has high processing efficiency and accuracy in handling short text translation. However, in the face of increasingly long texts and complex requirements, machine translation models composed of ML algorithms can no longer meet the needs of most scenarios[65].

**4. High Performance NLP Mode based on DCNN**

This article explored the feasibility of a high performance NLP model based on DCNN to address the complex pre work content, limited application scenarios, and difficulty in text annotation in traditional NLP models. This high performance NLP model integrates various technologies such as DL, CNN, and DCNN into multiple basic tasks in the NLP model, which to some extent improves the work efficiency and language processing accuracy of traditional NLP models[66].

In the face of modern complex text public datasets, this article deeply integrated CRF and CNN, and attempted to use CRF combined with CNN in the sentence analysis section to solve the sequence annotation problem. By adding a CRF processing layer on the bidirectional NN, the sequence annotation was carried out.

In this process, the back propagation (BP) algorithm in CNN was used, which can adjust the weights and parameters of multiple modules in the statement analysis model based on the error between the expected output value and the actual output value. This adjustment would continue until the actual output result of the model remains within a stable range. In this process, it is necessary to solve the error change rate of the model, as shown in the Formula (20) [13].

$$\delta = \frac{dE}{db} \quad (20)$$

Among them, E represents the error function. The backpropagation application of the BP algorithm is shown in Formula (21).

$$\delta^l = (W^{l+1})^T \delta^{l+1} \cdot f'(u^l) \quad (21)$$

The partial derivatives of the specified CNN layer and the weights of neurons in this layer have been updated, as shown in Formula (22).

$$\Delta W = -n \frac{dE}{dW^l} \quad (22)$$

The structured IE workflow in traditional NLP models is analyzed and optimized, mainly by combining DCNN with NER.

Firstly, the given sample data is preprocessed in the input layer and the sequence of each word in the text is determined. Then, the input sample data is transformed into the form of a word vector, which can be learned along with the training process of the model in NER.

The next layer is the convolutional layer of DCNN, which is mainly used in NER to extract local feature values in sentences. Combining DCNN with NER can quickly recognize and extract phrases or keywords in sentences. In this process, it is necessary to calculate the size of the feature images generated by the convolutional kernel, as shown in the Formula (23).

$$W' = \frac{(W+2p-k+s)}{s} \quad (23)$$

Next is the pooling layer of DCNN. The pooling layer generally reduces the feature dimensions of the sample data by performing maximum pooling operations on the data and preserves the most significant features of the data during this calculation process. In addition, the pooling operation of DCNN can also calculate the maximum value of local features in text data, thereby reducing the overall workload of NER[67].

After that, there is the stacking layer of DCNN, which stacks multiple convolutional and pooling layers, while also playing a role in building a deep level feature extraction model and improving the final expression ability of the model. Finally, there is the fully connected layer, which fuses and transforms the output feature values to enable the model to output the final result. In the NER task of the NLP model, the final output of the fully connected layer of DCNN is usually in the form of probability distribution, representing the possible entity relationships in the text.

Moreover, when integrating DCNN into NER to construct a new NER pattern, it is necessary to continuously adjust the model parameters through loss functions and backpropagation algorithms, which can further improve the accuracy of the model. The calculation formula for the loss function is shown in Formula (24) [14].

$$E = \frac{1}{2} \left\| d - f^i \right\|_l^2 \quad (24)$$

Among them, $f^i$ represents the final output in vector form, and the expected output value is represented by d. The next step is to calculate the total cost of training data, as shown in the Formula (25).

$$f(e) = \frac{1}{n} \sum_{i=1}^{n} e_i \quad (25)$$

During this process, it is necessary to continuously adjust the weight and bias values of the model to further reduce the overall error of the model calculation. At this point, it is necessary to solve for the minimum value of the overall cost, as shown in the Formula (26).

$$w_{ij} = w_{ij}^l - na^l \frac{dE}{dz^{(l+1)}} (26)$$

Finally, the error of the output result can be calculated using Formula (27).

$$\propto = (y - a_i)a^l(1 - a_i)(27)$$

At this point, the NER model based on DCNN can effectively analyze the entity relationships in the text sample data, and output the most accurate result through calculation.

In the task of text classification and text generation in the NLP model, this article attempted to integrate GAN into traditional NLP models, so that the NLP model has better multi-dimensional performance. In text classification and text generation tasks that combine GAN, the importance of vocabulary is first calculated and a set of words with the same meaning is constructed. Next, the characteristics of vocabulary are calculated and classified [15].

By integrating DCNN, GAN, ML, and soft computing into various basic tasks of traditional NLP models, the performance of multiple basic tasks of traditional NLP models is improved. In the text classification and text generation model that integrates GAN, the generation model first generates text based on the given data, and then delivers it to the recognition model to analyze the text generated by the generation model to determine the quality of the generated text. This process is repeated and the final high-quality text is output. The high performance NLP model obtained from this can also have better performance in more fields[68].

**5. High Performance NLP Model Validation and Experimentation**

This article analyzed the traditional NLP models commonly used in many fields. This analysis clarified that traditional NLP models have certain shortcomings in terms of sentence analysis accuracy and overall work efficiency, and therefore cannot adapt to the more demands of the times. Therefore, this article attempted to integrate various information technologies such as DCNN, GAN, and soft computing into multiple basic tasks of traditional NLP models, and referred to the new NLP model as a high performance NLP model. At the same time, it conducted empirical analysis on the performance of high performance NLP models and traditional NLP models in multiple aspects.

This article used multiple NLP common text datasets to construct four types of text test datasets during the experimental process, and their various parameters are shown in Table 2.

This article selected some text from the Movie review text test dataset and analyzed and processed the test set using the segmentation model from traditional NLP models and the segmentation model from high performance NLP models, respectively. The standard output word segmentation number of the Movie review text test dataset was 510 words, among which the traditional NLP model's word segmentation model output 604 word segmentation results, and only 435 results of the traditional NLP model's word segmentation model output meet the standard output word segmentation. The segmentation model of the high performance NLP model output 557 segmentation results, of which 454 segmentation results met the standard output segmentation. During this process, accuracy and recall were used to evaluate the segmentation performance of the two NLP models. The calculation formulas for accuracy and recall are shown in formulas (28) and (29).

$$P = \frac{n}{N} \times 100\% \quad (28)$$

$$R = \frac{n}{M} \times 100\% \quad (29)$$

Among them, n represents the number of words that the output results of different models meet the standard output segmentation; N represents the total number of word segments output by different segmentation models on the test dataset; M represents the number of standard output word segments. Based on the above formula, the word segmentation performance of the traditional NLP model and the high performance NLP model were calculated. During the calculation process, only two decimal places were retained before converting the results to percentages. The segmentation accuracy and recall of traditional NLP models were 72% and 85%, respectively, while the segmentation accuracy and recall of high performance NLP models were 82% and 89%, respectively. Thus, it was determined that the segmentation accuracy of the high performance NLP model improved by 10% and the recall rate increased by 4%.

In order to reduce the error of the experimental results, this article constructed 10 sets of text datasets containing 4 types of text by selecting partial test texts from 4 text test datasets. Two NLP model segmentation models were used to segment the 10 sets of text datasets, and their accuracy was recorded (the results were kept to two decimal places and not converted to percentages).

In Figure 6, the horizontal axis represents the traditional NLP segmentation model and the high performance NLP segmentation model, respectively; the vertical axis represents the segmentation accuracy of the two segmentation models. After analyzing Figure 6, it can be seen that when segmenting 10 sets of text test datasets, the segmentation accuracy of the high performance NLP model was distributed in the range of 0.74 to 0.9, while the segmentation accuracy of the traditional NLP model was distributed in the range of 0.6 to 0.85. The main reason for this accuracy analysis was the combination of high performance NLP models with various information technologies such as DL, CNN, and DCNN, which have better text processing capabilities. This text processing capability enables high performance NLP models to generally have higher word segmentation accuracy when segmenting text. The high performance NLP segmentation model had significantly higher segmentation accuracy.

Experimental analysis was conducted on the performance of two NLP models in text classification and text generation. During this process, ROUGE (Recall-Oriented Understudy for Gisting Evaluation) was used as a metric to measure the performance of two NLP models in text classification and text generation. ROUGE can evaluate the performance of model text classification and text generation by calculating the overlap between the model output results and the reference text. The experiment also used 10 sets of text test sets containing 4 different text test sets, and the calculation formula for the evaluation results of ROUGE is shown in Formula (30).

$$ROUGE - L = \frac{(1+B^2)RP}{R+B^2P} \quad (30)$$

According to Formula (30), the scores of two NLP text classification and text generation models can be statistically analyzed, and two histograms can be drawn based on two sets of data to display the scores of the two NLP text classification and text generation models, as shown in Figure 7.

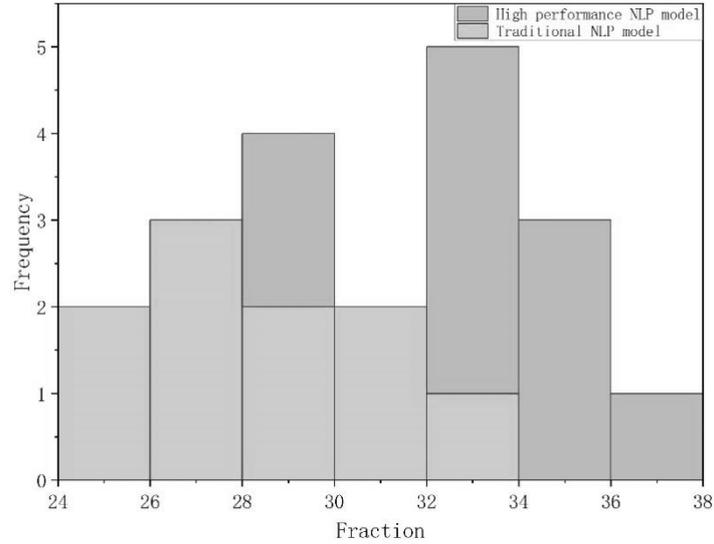

**Figure 7.** Presentation of scores for text classification and text generation using two NLP models

Figure 7 shows the ROUGE scores for text classification and text generation using traditional NLP models and high performance NLP models. The horizontal axis represents the score of the model, and the vertical axis represents the frequency of the model's score occurrence. In the distribution of scores between the two models, it was evident that the text classification and generation scores of the high performance NLP model were more concentrated above 30, while only two models with scores below 30. The average distribution of scores in the traditional NLP model was between 24 and 34, with only 3 scores greater than 30.

Experimental research was conducted on the performance of machine translation modules for two NLP models. In this process, the BLEU (Bilingual Evaluation Understudy) evaluation method was mainly used, which can automatically calculate the similarity between machine translated text and professional human translated text, and evaluate the quality of machine translation based on this value. In order to reduce errors in the experiment[67, 568], this article selected texts from the Wikipedia dataset for the experiment. The traditional NLP machine translation module and high performance NLP machine translation module were used to translate some texts from the Wikipedia dataset, and their scores were calculated using the BLEU evaluation method. The calculation method of BLEU is shown in Formula (31).

$$\text{BLEU} = \text{B} \cdot \text{ex}[\sum_{i=1}^{n} wlogp_n] (31)$$

Among them, $p_n$ represents the accuracy value; w represents a positive weight, and B represents a penalty factor. This article selected 100 short texts from Wikipedia datasets and divided them into 20 groups[69].

In Figure 8, the horizontal axis represents different NLP machine translation models, and the vertical axis represents the quality of two NLP machine translation models in translating 20 sets of short texts. After analyzing the scores of the NLP machine translation model in Figure 8, it can be determined that the scores of the high performance NLP machine translation model were mostly concentrated around 0.8, while the scores of the traditional NLP machine translation model were mostly concentrated around 0.75. The closer the score approaches 1 in the BLEU evaluation method, the better the translation performance of the machine translation model. The reason why the machine translation model of the high performance NLP model has better translation performance was still the integration of ML algorithm in the machine translation module, which can achieve higher accuracy in character matching work in machine translation[70].

Experimental evaluations were conducted on the overall operational efficiency of two NLP models. This evaluation mainly refers to the time required for different modules of two NLP models to handle the same task. In the experiment, different texts from four datasets were selected to form four types of text test sets. The efficiency of the four modules of sentence analysis, information extraction, text classification and generation, and machine translation of the two NLP models was analyzed. In the testing of a type of module, it is necessary to repeat it 10 times and collect the time used to draw a table. The time units used in the experiment are all seconds. The data is shown in figures 9, 10, 11, and 12.

After analyzing Figure 9, it can be determined that the average time spent on sentence analysis in the traditional NLP model in 10 experiments was 0.74 seconds, while the average time spent on sentence analysis in the high performance NLP model was 0.686 seconds. From this, it can be seen that the high performance NLP model had better performance in processing the same statements using various information technologies such as DL, CNN, and DCNN. It mainly relied on the algorithm models in DL, CNN, and DCNN to improve the analysis and processing capabilities of the three models of word analysis, syntactic analysis, and semantic analysis in sentence analysis, which improved the comprehensive ability of NLP's sentence analysis and ultimately improves the overall efficiency of sentence analysis.

After analyzing the information extraction time of the two models in Figure 10 for 10 experiments, it was determined that the average time of the traditional NLP model was 0.079 seconds, while the average time of the high performance NLP model was 0.058 seconds. The traditional NLP model consumed more time in each of the 10 information extraction experiments than the high performance NLP model, mainly because the combination of DCNN and NER enabled the NLP model to process more text information in the same time. This also indicated that the information extraction mode combined with DCNN and NER had better information extraction efficiency[71].

After studying the 10 experiments on text classification and text generation of the two models in Figure 11, it was found that the average time of the traditional NLP model was 1.624 seconds, while the average time of the high performance NLP model was 1.214 seconds. In the text classification and text generation experiments of the NLP model, the high performance NLP model can utilize GAN to continuously generate new text and deliver it to the discriminative model for judgment. This repetitive workflow enables the high performance NLP model to have better text classification capabilities, while also being able to output the most standard compliant text during the continuous adjustment process.

After calculating the 10 experiments of the two machine translation modules in Figure 12, it was determined that the average time used for the traditional NLP model was 2.366 seconds, while the average time used for the high-performance NLP model was 2.167 seconds. In the machine translation experiment of the NLP model, the machine translation module improved the machine translation data processing ability of the NLP model by integrating the ML algorithm model into string matching and vocabulary interpretation. This improvement in data processing ability improved the translation efficiency of the machine translation module.

Based on the analysis of the average time spent on 10 experiments of the traditional NLP model and the high-performance NLP model in figures 9, 10, 11, and 12, it was determined that high performance had better time performance on these four basic tasks[71].

## 6. Conclusion

In response to the low processing accuracy and operational efficiency of multiple basic tasks in traditional NLP models, this article introduced various information technologies such as DCNN and GAN into the multiple tasks in the traditional NLP model. This article explored the feasibility and reliability of a high-performance NLP model that combines multiple information technologies such as DCNN and GAN. It mainly evaluated the performance of high-performance NLP models and traditional NLP models through experiments, and set up four text test datasets to determine that the high-performance NLP model has better optimization effects in task processing accuracy and efficiency compared to traditional NLP models. Finally, this article verified the feasibility and reliability of a high-performance NLP that combines multiple information technologies such as DCNN and GAN. This model can provide a more accurate and efficient solution for NLP tasks.

## References

[1] Diksha Khurana, Aditya Koli, Kiran Khatter, and Sukhdev Singh. "Natural language processing: State of the art, current trends and challenges." Multimedia tools and applications 82.3 (2023): 3713-3744.

[2] Andrea, Galassi, Marco Lippi, and Paolo Torroni. "Attention in natural language processing." IEEE transactions on neural networks and learning systems 32.10 (2020): 4291-4308.

[3] XiPeng Qiu, TianXiang Sun, YiGe Xu, YunFan Shao, Ning Dai, and XuanJing Huang. "Pre-trained models for natural language processing: A survey." Science China Technological Sciences 63.10 (2020): 1872-1897.

[4] Dastan Hussen Maulud, Subhi R. M. Zeebaree, Karwan Jacksi, Mohammed A. Mohammed Sadeeq, and Karzan Hussein Sharif. "State of art for semantic analysis of natural language processing." Qubahan academic journal 1.2 (2021): 21-28.

[5] Yu Gu, Robert Tinn, Hao Cheng, Michael Lucas, Naoto Usuyama, Xiaodong Liu, et al. "Domain-specific language model pretraining for biomedical natural language processing." ACM Transactions on Computing for Healthcare (HEALTH) 3.1 (2021): 1-23.

[6] Pengfei Liu, Weizhe Yuan, Jinlan Fu, Zhengbao Jiang, Hiroaki Hayashi, and Graham Neubig. "Pre-train, prompt, and predict: A systematic survey of prompting methods in natural language processing." ACM Computing Surveys 55.9 (2023): 1-35.

[7] Ergin Soysal, Jingqi Wang, Min Jiang, Yonghui Wu, Serguei Pakhomov, Hongfang Liu, et al. "CLAMP–a toolkit for efficiently building customized clinical natural language processing pipelines." Journal of the American Medical Informatics Association 25.3 (2018): 331-336.

[8] Daniel W., Otter, Julian R. Medina, and Jugal K. Kalita. "A survey of the usages of deep learning for natural language processing." IEEE transactions on neural networks and learning systems 32.2 (2020): 604-624.

[9] Tom Young, Devamanyu Hazarika, Soujanya Poria, and Erik Cambria. "Recent trends in deep learning based natural language processing." ieee Computational intelligenCe magazine 13.3 (2018): 55-75.

[10] Jalal Rezaeenour, Mahnaz Ahmadi, Hamed Jelodar, and Roshan Shahrooei. "Systematic review of content analysis algorithms based on deep neural networks." Multimedia Tools and Applications 82.12 (2023): 17879-17903.

[11] Haitao Wang, Keke Tian, Zhengjiang Wu, and Lei Wang. "A short text classification method

based on convolutional neural network and semantic extension." International Journal of Computational Intelligence Systems 14.1 (2021): 367-375.

[12]   Yixing Li, Zichuan Liu, Wenye Liu, Yu Jiang, Yongliang Wang, Wang Ling Goh, et al. "A 34-FPS 698-GOP/s/W binarized deep neural network-based natural scene text interpretation accelerator for mobile edge computing." IEEE Transactions on Industrial Electronics 66.9 (2018): 7407-7416.

[13]   Dr Jennifer S., Raj, and J. Vijitha Ananthi. "Recurrent neural networks and nonlinear prediction in support vector machines." Journal of Soft Computing Paradigm 1.1 (2019): 33-40.

[14]   Ding-Xuan, Zhou. "Universality of deep convolutional neural networks." Applied and computational harmonic analysis 48.2 (2020): 787-794.

[15]   Jie Gui, Zhenan Sun, Yonggang Wen, Dacheng Tao, and Jieping Ye. "A review on generative adversarial networks: Algorithms, theory, and applications." IEEE transactions on knowledge and data engineering 35.4 (2021): 3313-3332.

[16]   Tom Young, Devamanyu Hazarika, Soujanya Poria, and Erik Cambria. "Recent trends in deep learning based natural language processing." ieee Computational intelligenCe magazine 13.3 (2018): 55-75.

[17]   Tom Young, Devamanyu Hazarika, Soujanya Poria, and Erik Cambria. "Recent trends in deep learning based natural language processing." ieee Computational intelligenCe magazine 13.3 (2018): 55-75.

[18]   Ergin Soysal, Jingqi Wang, Min Jiang, Yonghui Wu, Serguei Pakhomov, Hongfang Liu, et al. "CLAMP–a toolkit for efficiently building customized clinical natural language processing pipelines." Journal of the American Medical Informatics Association 25.3 (2018): 331-336.

[19]   Daniel W., Otter, Julian R. Medina, and Jugal K. Kalita. "A survey of the usages of deep learning for natural language processing." IEEE transactions on neural networks and learning systems 32.2 (2020): 604-624.

[20]   Tom Young, Devamanyu Hazarika, Soujanya Poria, and Erik Cambria. "Recent trends in deep learning based natural language processing." ieee Computational intelligenCe magazine 13.3 (2018): 55-75.

[21]   Khodaverdian, Zeinab, et al. "Combination of convolutional neural network and gated recurrent unit for energy aware resource allocation." arXiv preprint arXiv:2106.12178 (2021).X. Gao, C. Gupta, and H. Li, "Automatic Lyrics Transcription of Polyphonic Music With Lyrics-Chord Multi-Task Learning," in IEEE/ACM Transactions on Audio, Speech, and Language Processing, vol. 30, pp. 2280-2294, 2022.

[22]   Nazari, M., Moayed Rezaie, S., Yaseri, F. et al. Design and analysis of a telemonitoring system for high-risk pregnant women in need of special care or attention. BMC Pregnancy Childbirth 24, 817 (2024).

[23]   Sadr, H., et al., Cardiovascular disease diagnosis: a holistic approach using the integration of machine learning and deep learning models. European Journal of Medical Research, 2024. 29(1): p. 455.

[24]   Yu, Y., Tang, S., Raposo, F., and Chen, L., "Deep cross-modal correlation learning for audio and lyrics in music retrieval", ACM Transactions on Multimedia Computing, Communications, and Applications (TOMM), vol.15(1), pp.1-16, 2019.

[25] Tahbaz, Mahdi, Hossein Shirgahi, and Mohammad Reza Yamaghani. "Evolutionary-based image encryption using Magic Square Chaotic algorithm and RNA codons truth table." Multimedia Tools and Applications 83.1 (2024): 503-526.

[26] Oramas, S., Barbieri, F., Nieto Caballero, O. and Serra, X., "Multimodal deep learning for music genre classification", Transactions of the International Society for Music Information Retrieval, vol.1, pp.4-21, 2018.

[27] Delbouys, R., Hennequin, R., Piccoli, F., Royo-Letelier, J. and Moussallam, M., "Music mood detection based on audio and lyrics with deep neural net", 2018.

[28] Chen, C. and Li, Q., "A multimodal music emotion classification method based on multifeature combined network classifier", Mathematical Problems in Engineering, vol. 1, pp.4606027, 2020.

[29] Keshavarz, Hadis, and Hossein Sadr. "Object Detection for Automated Coronary Artery Using Deep Learning." arXiv preprint arXiv:2312.12135 (2023).

[30] Khobdeh, Soroush Babaee, Mohammad Reza Yamaghani, and Siavash Khodaparast Sareshkeh. "Basketball action recognition based on the combination of YOLO and a deep fuzzy LSTM network." The Journal of Supercomputing 80.3 (2024): 3528-3553.

[31] Parisi, L., Francia, S., Olivastri, S. and Tavella, M.S., "Exploiting synchronized lyrics and vocal features for music emotion detection", Computation and Language, 2019.

[32] Sadr, H., et al., A shallow convolutional neural network for cerebral neoplasm detection from magnetic resonance imaging. Big Data and Computing Visions, 2024. 4(2): p. 95-109.

[33] Durmus Ozkan Sahin, Oguz Emre Kural, Erdal Kilic, Armagan Karabina, "A Text Classification Application: Poet Detection from Poetry", Information Retrieval, 2018.

[34] Tahbaz, Mahdi, Hossein Shirgahi, and Mohammad Reza Yamaghani. "Evolutionary-based image encryption using Magic Square Chaotic algorithm and RNA codons truth table." Multimedia Tools and Applications 83.1 (2024): 503-526.

[35] Khodaverdian, Z., et al., An energy aware resource allocation based on combination of CNN and GRU for virtual machine selection. Multimedia tools and applications, 2024. 83(9): p. 25769-25796.

[36] Saberi, Z.A., H. Sadr, and M.R. Yamaghani. An Intelligent Diagnosis System for Predicting Coronary Heart Disease. in 2024 10th International Conference on Artificial Intelligence and Robotics (QICAR). 2024. IEEE.

[37] Zahmatkesh Zakariaee, A., H. Sadr, and M.R. Yamaghani, A new hybrid method to detect risk of gastric cancer using machine learning techniques. Journal of AI and Data Mining, 2023. 11(4): p. 505-515.

[38] Khodaverdian, Z., et al., Predicting the workload of virtual machines in order to reduce energy consumption in cloud data centers using the combination of deep learning models. Journal of Information and Communication Technology, 2023. 55(55): p. 166.

[39] Sadr, H., et al., Virtual Machine Workload Prediction to Reduce Energy Consumption in Cloud Data Centers Using Combination of Deep Learning Models. Journal of Information and Communication Technology, 2023. 55(56): p. 1.

[40] Mohades Deilami, F., H. Sadr, and M. Tarkhan, Contextualized multidimensional personality recognition using combination of deep neural network and ensemble learning. Neural Processing Letters, 2022. 54(5): p. 3811-3828.

[41]  Hosseini, Seyed Sadegh, Mohammad Reza Yamaghani, and Soodabeh Poorzaker Arabani. "Multimodal modelling of human emotion using sound, image and text fusion." Signal, Image and Video Processing 18.1 (2024): 71-79.

[42]  Sadr, H. and M. Nazari Soleimandarabi, ACNN-TL: attention-based convolutional neural network coupling with transfer learning and contextualized word representation for enhancing the performance of sentiment classification. The Journal of Supercomputing, 2022. 78(7): p. 10149-10175.

[43]  Deilami, Fatemeh Mohades, Hossein Sadr, and Mojdeh Nazari. "Using machine learning based models for personality recognition." arXiv preprint arXiv:2201.06248 (2022).

[44]  Babaee Khobdeh, S., M. R. Yamaghani, and S. Khodaparast. "A novel method for clustering basketball players with data mining and hierarchical algorithm." International Journal of Applied Operational Research-An Open Access Journal 11.4 (2023): 25-38.

[45]  Mirhoseini-Moghaddam, Seyedeh Mahsa, Mohammad Reza Yamaghani, and Adel Bakhshipour. "Application of electronic nose and eye systems for detection of adulteration in olive oil based on chemometrics and optimization approaches." Journal of Universal Computer Science 29.4 (2023): 300.

[46]  Bodó, Z. and Szilágyi, E., "Connecting the last. fm dataset to lyricwiki and musicbrainz lyrics-based experiments in genre classification", vol. 10(2), pp.158-182, 2018.

[47]  Langeroudi, Milad Keshtkar, Mohammad Reza Yamaghani, and Siavash Khodaparast. "FD-LSTM: A fuzzy LSTM model for chaotic time-series prediction." IEEE Intelligent Systems 37.4 (2022): 70-78.

[48]  Shayan Rostami, Hossein Sadr, Seyed Ahmad Edalatpanah, Mojdeh Nazari, "IQ Estimation from MRI Images using a Combination of Convolutional Neural Network and XGBoost algorithm", IEEE Access, vol. 8, pp. 73865-73878, 2020.

[49]  Xiaoguang Jia, "Music Emotion Classification Method Based on Deep Learning and Improved Attention Mechanism", Computational intelligence and neuroscience, 20 June 2022.

[50]  Curtis Thompson, "Lyric-Based Classification of Music Genres Using Hand-Crafted Features", An International journal of undergraduate research, Vol. 14, No. 2, 2021.

[51]  Sadr, Hossein. "An intelligent model for multidimensional personality recognition of users using deep learning methods." Journal of Information and Communication Technology 47.47 (2021): 72.

[52]  Hosseini, Seyed Sadegh, Mohammadreza Yamaghani, and Soodabeh Poorzaker Arabani. "A Review of Methods for Detecting Multimodal Emotions in Sound, Image and Text." Journal of Applied Research on Industrial Engineering (2024).

[53]  Sadr, H., M.M. Pedram, and M. Teshnehlab, Multi-view deep network: a deep model based on learning features from heterogeneous neural networks for sentiment analysis. IEEE access, 2020. 8: p. 86984-86997.

[54]  Sadr, H., M.M. Pedram, and M. Teshnehlab, Convolutional neural network equipped with attention mechanism and transfer learning for enhancing performance of sentiment analysis. Journal of AI and data mining, 2021. 9(2): p. 141-151.

[55]  Soleymanpour, S., H. Sadr, and M. Nazari Soleimandarabi, CSCNN: cost-sensitive convolutional neural network for encrypted traffic classification. Neural Processing Letters, 2021. 53(5): p. 3497-3523.

[56] Kalashami, M.P., M.M. Pedram, and H. Sadr, EEG feature extraction and data augmentation in emotion recognition. Computational intelligence and neuroscience, 2022. 2022(1): p. 7028517.

[57] Nazari, M., et al., Detection of Cardiovascular Diseases Using Data Mining Approaches: Application of an Ensemble-Based Model. Cognitive Computation, 2024: p. 1-15

[58] Soleimandarabi, Mojdeh Nazari, Seyed Abolghasem Mirroshandel, and Hossein Sadr. "The Significance of Semantic Relatedness and Similarity measures in Geographic Information Science." International Journal of Computer Science and Network Solutions 3.2 (2015): 12-23.

[59] Z. Jin, X. Lai, and J. Cao, "Multi-label Sentiment Analysis Base on BERT with modified TF-IDF," Product Compliance Engineering-Asia, pp. 1-6, 2020.

[60] Sajjad Amiri Doumari, Hadi Givi, Mohammad Dehghani, Om Parkash Malik, "Ring Toss Game-Based Optimization Algorithm for Solving Various Optimization Problems," International Journal of Intelligent Engineering and Systems, vol. 14, Issue. 3, pp. 545-554, 2021.

[61] Dehghan, S., et al., Comparative study of machine learning approaches integrated with genetic algorithm for IVF success prediction. Plos one, 2024. 19(10): p. e0310829.

[62] Ma Y., Xu Y., Liu Y., Yan F., Zhang Q., Li Q. and Liu Q., "Multi-Scale Cross-Attention Fusion Network Based on Image Super-Resolution" Applied Sciences, vol. 14(6), pp.2634, 2024.

[63] Adel Sabry Eesa, Zeynep Orman, Adnan Mohsin Abdulazeez Brifcani, "A novel feature-selection approach based on the cuttlefish optimization algorithm for intrusion detection systems", Expert Systems with Applications, Vol. 42, Issue 5, pp. 2670-2679, 1 April 2015.

[64] Mohammad Dehghani and Pavel Trojovský, "Osprey optimization algorithm: A new bio-inspired metaheuristic algorithm for solving engineering optimization problems", Frontiers in Mechanical Engineering, Vol. 8, 2022.

[65] Jalal Rezaeenour, Mahnaz Ahmadi, Hamed Jelodar, and Roshan Shahrooei. "Systematic review of content analysis algorithms based on deep neural networks." Multimedia Tools and Applications 82.12 (2023): 17879-17903.

[66] Langeroudi, Milad Keshtkar, Mohammad Reza Yamaghani, and Siavash Khodaparast. "FD-LSTM: A fuzzy LSTM model for chaotic time-series prediction." IEEE Intelligent Systems 37.4 (2022): 70-78.

[67] Haitao Wang, Keke Tian, Zhengjiang Wu, and Lei Wang. "A short text classification method based on convolutional neural network and semantic extension." International Journal of Computational Intelligence Systems 14.1 (2021): 367-375.

[68] Khodaverdian, Zeinab, Hossein Sadr, and Seyed Ahmad Edalatpanah. "A shallow deep neural network for selection of migration candidate virtual machines to reduce energy consumption." 2021 7th International conference on web research (ICWR). IEEE, 2021.

[69] Yixing Li, Zichuan Liu, Wenye Liu, Yu Jiang, Yongliang Wang, Wang Ling Goh, et al. "A 34-FPS 698-GOP/s/W binarized deep neural network-based natural scene text interpretation accelerator for mobile edge computing." IEEE Transactions on Industrial Electronics 66.9 (2018): 7407-7416.

[70] Hossein Sadr, Reza Ebrahimi Atani, Mohammad Reza Yamaghani, " The Significance of Normalization Factor of Documents to Enhance the Quality of Search in Information Retrieval Systems" International Journal of Computer Science and Network Solutions 2.5 (2014): 91-97.

[71] Jennifer S., Raj, and J. Vijitha Ananthi. "Recurrent neural networks and nonlinear prediction in support vector machines." Journal of Soft Computing Paradigm 1.1 (2019): 33-40.